\documentclass{article}
\usepackage{spconf,amsmath,graphicx,hyperref,amssymb,booktabs,multirow, cite}
\usepackage[table]{xcolor}
\usepackage{etoolbox}
\usepackage{enumitem}
\usepackage{multirow,makecell}
\apptocmd{\thebibliography}{\setlength{\itemsep}{0pt}\setlength{\parskip}{0pt}}{}{}
\title{Zero-Shot VISUAL GROUNDING in 3D Gaussians via View Retrieval}
\address{Author Affiliation(s)}
\name{Liwei Liao\textsuperscript{\rm 1,2}, 
  Xufeng Li\textsuperscript{\rm 3},
  Xiaoyun Zheng\textsuperscript{\rm 1,2},
  Boning Liu\textsuperscript{\rm 2},
  Feng Gao\textsuperscript{\rm 1}\sthanks{Corresponding Author},
  Ronggang Wang\textsuperscript{\rm 2}}
\address{
\textsuperscript{\rm 1} Peking University,
\textsuperscript{\rm 2} Peng Cheng Laboratory, 
\textsuperscript{\rm 3} City University of Hongkong
}

\begin{document}
%\ninept
%
\maketitle
\begin{abstract}
3D Visual Grounding (3DVG) aims to locate objects in 3D scenes based on text prompts, which is essential for applications such as robotics. However, existing 3DVG methods encounter two main challenges: first, they struggle to handle the implicit representation of spatial textures in 3D Gaussian Splatting (3DGS), making per-scene training indispensable; second, they typically require larges amounts of labeled data for effective training. To this end, we propose \underline{G}rounding via \underline{V}iew \underline{R}etrieval (GVR), a novel zero-shot visual grounding framework for 3DGS to transform 3DVG as a 2D retrieval task that leverages object-level view retrieval to collect grounding clues from multiple views, which not only avoids the costly process of 3D annotation, but also eliminates the need for per-scene training. Extensive experiments demonstrate that our method achieves state-of-the-art visual grounding performance while avoiding per-scene training, providing a solid foundation for zero-shot 3DVG research. Video demos can be found in \url{https://github.com/leviome/GVR_demos}.

% 3D Visual Grounding (3DVG) aims to locate objects in 3D scenes based on textual descriptions, essential for applications like augmented reality and robotics.
% 3DVG是通过语言提示实现三维场景的目标定位，这对于增强现实和机器人等应用至关重要。
% 然而, 现有的3DVG方法主要处理网格或点云，无法处理另一种高效的三维表示技术3DGS, 因为3DGS的空间纹理是隐式表示,导致其不容易被神经网络聚合特征。
\end{abstract}
\begin{keywords}
3D Visual Grounding, View Retrieval, Scene Understanding
\end{keywords}
\section{Introduction}
\label{sec:intro}
% 近年来，随着语言-视觉大模型的发展，3D视觉定位迎来了一波发展。由于三维场景的数字表示并没有达到统一（比如Mesh，点云等），因此针对不同的三维表示会有对应的3DVG方案。3D高斯是一种很有潜力的三维表示方式，然而由于其空间纹理是隐式表示，因此现有的3DVG方法并不适用于3D高斯。为此，我们提出了一种基于视图检索的3D高斯视觉定位方法，将3DVG任务转化为2D检索任务，从而实现对3D高斯场景的视觉定位。
3D Gaussian Splatting (3DGS)~\cite{kerbl20233d} offers significant advantages over traditional 3D reconstruction methods in terms of reconstruction speed, quality, and rendering efficiency. However, because 3DGS adopts a semi-implicit representation with explicit geometric structures and implicit spatial textures, it is difficult to directly apply previous point-cloud or radiance-field methods to 3DGS. 
% In recent years, with the development of vision-language large models~\cite{radford2021learning,ouyang2022training,achiam2023gpt,wang2024qwen2}, 3D visual grounding has experienced significant advancements~\cite{kerr2023lerf, li2025seeground, xu2024vlm,jain2025unifying,li2023voxformer,qin2024langsplat,yuan2024visual,liu2025reasongrounder}.
% Since the digital representations of 3D scenes remain diverse—including voxels, meshes, point clouds, NeRFs~\cite{mildenhall2021nerf}, and 3DGS~\cite{kerbl20233d}—different 3DVG solutions are tailored to specific representation formats. 
% 对于voxels, meshes和point cloud这种显示表示，通常可以训练一个端到端的模型来实现 3DVG任务，比如voxformer和UniVLG，也可以用导航的思路来做3DVG，比如SeeGround。对于隐式表示比如NeRF，则需要训练隐式的神经网络来实现language understanding，避免不了巨大的训练开销。然而对于半隐式表示比如 3DGS，其有着显示的几何结构和隐式的空间纹理，目前无法使用端到端模型来进行特征聚合，所以现有方法如LangSplat对逐场景训练出高斯级别的语义特征，以此完成3DVG。  
% 3DGS在重建速度、重建质量和渲染速度等方面，相比传统的三维重建方法都具有很大的优势。近年来，针对 3DGS订制场景理解算法的工作越来越多。但目前并没有很好的可泛化的针对 3DGS的 3DVG算法。

In recent years, there has been a growing body of work on developing scene understanding algorithms~\cite{GauGroup,choi2024click,gaussianeditor,SAGA,SAGD,shen2024flashsplat,zhu2025rethinking, liu2025reasongrounder,li2025scenesplat} specifically for 3DGS. These approaches can be categorized into two main types: (1) supervised 3DVG methods such as SceneSplat~\cite{li2025scenesplat} that require annotated 3D data for training and perform end-to-end inference, and (2) per-scene training 3DVG methods such as LangSplat~\cite{qin2024langsplat} that train a language Gaussian field for each specific scene. However, both the two approaches face different limitations: (1) For SceneSplat, pre-training is conducted only on 7,000 indoor scenes, so its generalization ability is limited to several indoor environments. To expand its generalization requires a massive amount of 3D annotated data, which is currently expensive. (2) For LangSplat, hour-scale per-scene preparation (including preprocessing and training) are required, and the query stage is also computationally complex. These two limitations severely restrict the application of 3DVG in the 3DGS domain, especially the requirement for hours of per-scene training, which greatly undermines its practicality.
% 然而这两种方法都面临不同的限制。对于SceneSplat来说，只是在 7000 个室内场景上进行预训练，导致它的可泛化性只集中在室内环境。如果需要扩大泛化性，则需增加海量的三维标注数据，但目前来讲，这种做法极不经济。对于Langsplat来说，需要小时级别的逐场景准备时间（包括预处理和训练），并且在Query阶段有较高的复杂度。
% 由于这两种限制，现在的方法
% 以LangSplat为代表的方法会通过一个编解码器将CLIP的二维语义特征蒸馏到三维高斯中，从而实现对三维高斯的视觉定位。但是LangSplat具有三个明显缺陷：1，需要针对具体场景进行小时级别的逐场景训练；2，有较低质量visual grounding效果；3，有高昂的显存和计算成本。
% 3DGS-based 3DVG methods such as LangSplat~\cite{qin2024langsplat} employ an encoder-decoder architecture to distill CLIP~\cite{radford2021learning}’s 2D semantic features into 3D Gaussians, thereby enabling visual grounding in 3D Gaussian scenes. However, LangSplat has three notable limitations: (1) it requires hours of per-scene training for each specific environment; (2) it delivers relatively low-quality visual grounding results; and (3) it incurs high memory and computational costs. These limitations severely restrict the application of 3DVG in the 3DGS domain, especially the requirement for hours of per-scene training, which greatly undermines its practicality.
% 这些缺点都严重限制了 3DVG在3DGS领域的应用，尤其是小时级别的逐场景训练，极大地影响了实用性。

% 对于第一个限制，我们采用现有成熟的2D visual foundation model来完成 3DVG的工作，极大程度地保留了这些 2D模型的可泛化性，本方法是完全training-free的。
% 对于第二个限制，我们采用了知识库的方式代替逐场景训练，不仅大大节省了逐场景准备时间，也加速了查询。

To address the two limitations, we propose a retrieval-based 3D Gaussian visual grounding method that reformulates the 3DVG task as a 2D retrieval problem, enabling effective visual grounding in 3D Gaussian scenes. For the first limitation, our work is training-free by leveraging existing mature 2D visual foundation models to accomplish visual perception, thereby largely preserving the generalization ability of these 2D models. For the second limitation, we replace per-scene training with a knowledge book approach, which not only greatly reduces preparation time for each scene but also accelerates the query process. Our main contributions are summarized as follows:
\begin{itemize}
  \item We introduce \textbf{GVR}, to the best of our knowledge, the first framework capable of zero-shot visual grounding in 3D Gaussian scenes.
  \item We propose a view retrieval mechanism for 3DVG, which eliminates the need for 3D annotated data and enables high-quality visual grounding using only existing mature 2D visual foundation models.
  % 我们提出了基于 2D视图检索的 3DVG 方案，避免了对 3D标注数据的需求，仅用现有成熟的 2D visual foundation model即可完成高质量的视觉定位。

  \item Extensive experiments demonstrate that our approach achieves state-of-the-art performance in zero-shot 3D visual grounding of 3DGS, while significantly saving training time.
\end{itemize}

% 为了解决以上问题，我们提出了一种基于视图检索的3D高斯视觉定位方法，将3DVG任务转化为2D检索任务，从而实现对3D高斯场景的视觉定位。我们的贡献总结如下：

% \section{Related Works}
% \label{sec:related works}

% Recent works in 3D visual grounding have explored various representations, including point clouds, meshes, and neural radiance fields. However, few methods are tailored for 3D Gaussian Splatting (3DGS), which offers efficient and high-quality scene modeling. Conventional approaches often require annotated datasets and supervised training, limiting their scalability and generalization.

% Zero-shot visual grounding aims to localize objects or regions in 3D scenes without explicit supervision. Existing zero-shot methods typically leverage pre-trained vision-language models or multi-view retrieval, but their adaptation to 3DGS remains underexplored. Our work builds upon multi-view image retrieval techniques, extending them to the context of 3D Gaussian representations.

% Compared to prior supervised methods, our approach does not rely on labeled data and can generalize to arbitrary queries and scenes. This enables practical deployment in real-world scenarios where annotated 3D data is scarce. We demonstrate that view retrieval in 3DGS provides a robust and scalable solution for zero-shot visual grounding, outperforming traditional baselines on diverse benchmarks.

\section{Methodology}
\label{sec:method}
% 我们的方法解决了一个面向 3DGS的零样本三维视觉定位问题。
Our method aims to address the problem of zero-shot 3D visual grounding specifically for 3DGS. The overall process can be formulated as:
\begin{equation}
\textbf{I}_{tar} = \operatorname{GVR}(Q;\textbf{G})
\end{equation}
where $\textbf{I}_{tar} \in \mathbb{R}^{N \times 1}$ indicates the flag of all Gaussians that 1 denotes target and 0 denotes background, $Q$ is the text query (e.g., a red apple) describing the desired object or region, $\textbf{G} \in \mathbb{R}^{N \times 59}$ denotes the 3D Gaussian scene containing $N$ Gaussians, and $\text{GVR}(\cdot)$ refers to our proposed 3DVG method.

% Our framework consists of three main components: (1) view retrieval, (2) multi-view stereo voting, and (3) frustum intersection. Given a visual query, we first retrieve the most relevant camera views from the multi-view images associated with the 3D Gaussian scene. Next, each selected view proposes candidate 3D locations for the query, which are aggregated using a voting mechanism to reinforce consistent predictions. Finally, we refine the localization by intersecting the camera frustums of the retrieved views, focusing on overlapping regions in the 3D Gaussian representation. This pipeline enables accurate zero-shot visual grounding in complex scenes without requiring annotated training data.

\begin{figure*}[htb]
  \centering
  \includegraphics[width=0.95\textwidth]{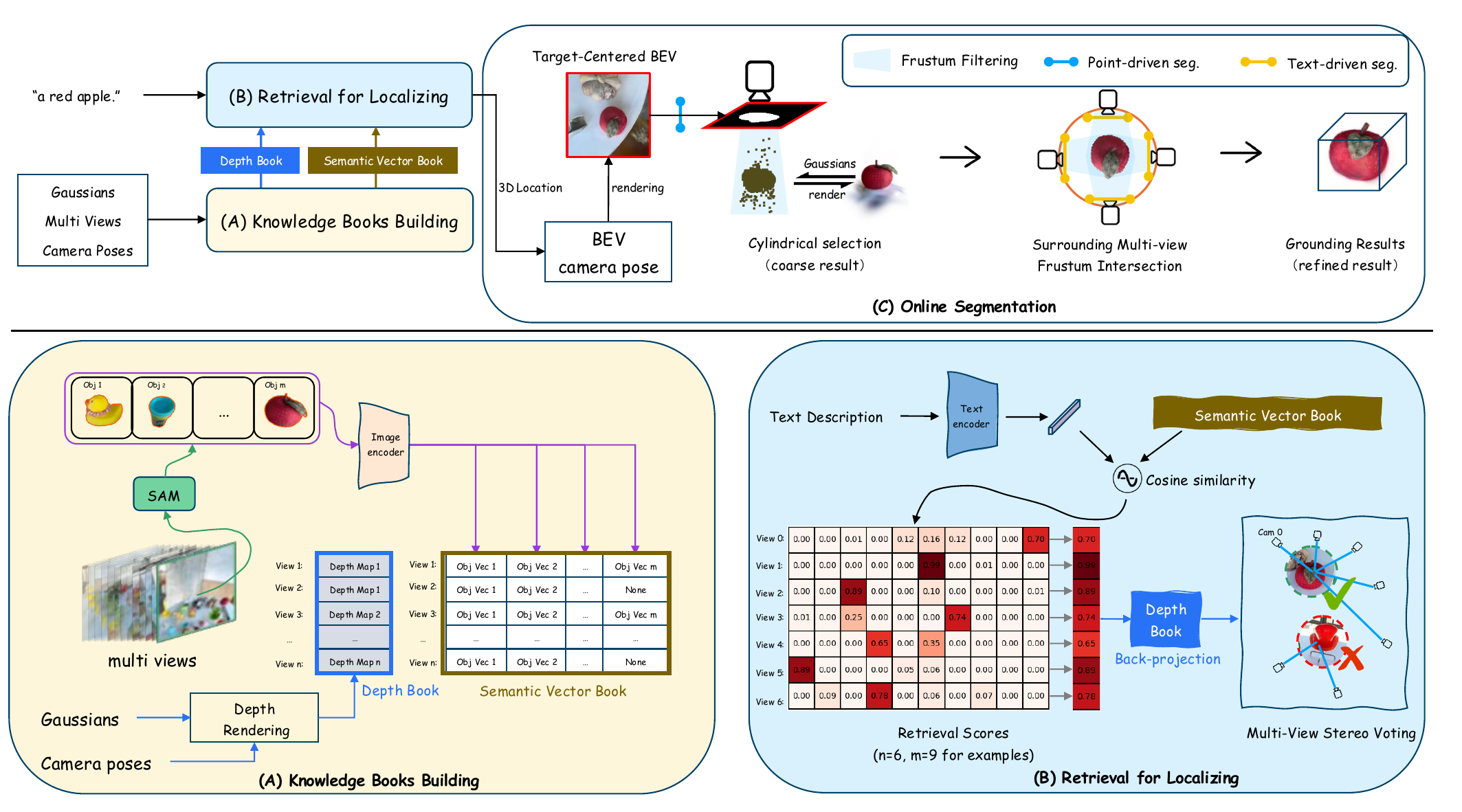}
  \vspace{-0.1in}
  \caption{\textbf{Overall pipeline of our proposed GVR}. (A) \textbf{Preparation}: Construct a Semantic Vector Book using SAM and CLIP, and obtain a Depth Book via 3DGS depth rendering. (B) \textbf{Query}: Encode the text query with CLIP’s text encoder, compute similarity with the multi-view semantic vector library, select top-k relevant views and object patches, and estimate the 3D object location via multi-view back-projection and stereo voting. (C) \textbf{Segmentation}: Perform point-driven segmentation on the BEV using the projected 3D location, obtain coarse target Gaussians via frustum filtering, and refine them through surrounding multi-view frustum intersection.
  % 在准备阶段（子图A）：通过SAM和CLIP构建成一个Semantic Vector Book，同时通过3DGS的深度渲染获得Depth Book。在Query阶段（子图B）：通过CLIP的text encoder将文本查询编码成语义向量，并与多视图的语义向量库进行相似度计算，选出top-k相关视图的相关物体patch，通过多视角下的反投影得到物体的三维位置。在Segmentation阶段(子图C)：通过3D Location的投影点在BEV上进行point-driven segmentation，通过Frustum Filtering得到粗略的target Gaussians，然后通过Surrounding Multi-view Frustum Intersection得到精细的target Gaussians。
  }
  \label{fig:framework}
  \vspace{-0.1in}
\end{figure*}

% \subsection{Preliminary: 3D Gaussian Splatting}
% \label{ssec:preliminary}

% By capturing $n$ views $\mathbf{V} = \left\{ v_0, v_1, \ldots, v_{n-1} \right\}$ from $n$ cameras, the camera poses $\mathbf{C} = \left\{ c_0, c_1, \ldots, c_{n-1} \right\}$ and a sparse point cloud of the scene can be reconstructed using Structure from Motion (SfM)~\cite{schonberger2016structure}. Starting from the sparse point cloud, 3DGS transforms the points into ellipsoidal Gaussians. Through differentiable rasterization, the supervision gradients from the original images to the rendered views can be backpropagated to the Gaussian primitives, enabling dynamic adjustment of their shape, orientation, spherical harmonics coefficients, opacity, and quantity, thereby accomplishing 3D reconstruction and representation. Ultimately, this process yields a 3D scene $\textbf{G} \in \mathbb{R}^{N \times 59}$ represented by $N$ Gaussian primitives.
% % 从稀疏点云开始，3DGS将它们椭球化，然后通过differentiable rasterization建立原始图对渲染视图的监督梯度能够传递到高斯基元，使高斯基元的形状、朝向、球谐系数、透明度和数量动态调整，从而完成3D重建和表示。最终形成N个高斯基元表示的三维场景。
% % 在3DVG任务中，模型通过文本描述从这N个高斯基元中找出属于目标物体的高斯基元，并用索引的方式进行表示。
% In the 3DVG task, the model identifies the Gaussian primitives corresponding to the target object from the $N$ candidates based on the textual description, and represents them by their indices $\textbf{I}_{target}$.

\subsection{(A) Preparation: Knowledge Books Building}
\label{ssec:preparation}
As shown in Fig.~\ref{fig:framework} (A), we first prepare Semantic Vector Book (SVB) and Depth Book (DB) $\{\textbf{B}_{s}, \textbf{B}_{d}\}$.
Since the 3D scene $\textbf{G}$ is reconstructed from multi-view images and camera parameters $\{\mathbf{V}, \mathbf{C}\}$, all semantic information about the scene can be inferred from these multi-view images. Based on this observation, we construct a SVB $\textbf{B}_{s} \in \mathbb{R}^{n \times m \times c}$ to store the scene’s semantic information. Specifically, we use SAM~\cite{kirillov2023segment} to segment each view and obtain multiple object masks, then encode each object patch into a semantic vector using CLIP’s~\cite{radford2021learning} image encoder, as:
  \begin{equation}
    \textbf{B}^{(i)(j)}_{s} = \mathcal{E}_{img}(\operatorname{SAM}(v_i)^{(j)})
  \end{equation}
where $\operatorname{SAM}(\cdot)$ denotes the SAM model that segments the input image into $m$ object patches, $\mathcal{E}_{img}(\cdot)$ represents CLIP’s image encoder that encodes each object patch into a $c$-dimensional semantic vector (typically $c=512$), and $\textbf{B}^{(i)(j)}_{s} \in \mathbb{R}^{c}$ is the semantic vector for the $j$-th object in the $i$-th view.
Meanwhile, we obtain a depth map for each view via 3DGS depth rendering, and aggregate these to build a DB $\textbf{B}_{d} \in \mathbb{R}^{n \times H \times W}$ for storing the scene’s depth information. This step can be formulated as:
  \begin{equation}
    \textbf{B}^{(i)}_{d} = \mathcal{D}(c_i;\textbf{G})
  \end{equation}
where $\mathcal{D}(\cdot;\textbf{G})$ denotes the depth rendering of the 3D Gaussian scene $\textbf{G}$, $c_i$ represents the camera parameters for the $i$-th view, and $\textbf{B}^{(i)}_{d} \in \mathbb{R}^{H \times W}$ is the depth map for the $i$-th view, which has the same resolution to view $v_i$.

% 由于3D scene是由多视角的视图和相机位姿重建得到，所以三维场景的所有语义信息都可以在多视角视图中找到线索。基于这个判断，我们决定构建一个多视图的语义向量库，来存储场景的语义信息。具体来说，我们使用SAM对每个视图进行分割，得到该视图的多个物体mask，然后通过CLIP的image encoder将每个物体mask对应的图像patch编码成一个语义向量。最终，我们将所有视图的语义向量进行汇总，形成一个Semantic Vector Book，来存储场景的语义信息。同时，我们通过3DGS的深度渲染，得到每个视图对应的深度图，从而构建一个Depth Book，来存储场景的深度信息。

\subsection{(B) Query: Retrieval For Localizing (RFL)}
\label{ssec:query}
As shown in Fig.~\ref{fig:framework} (B), this stage is for retrieving relevant patches of views for a given textual query and localizing objects within those clues. First, we encode the text query $Q$ (such as ``a red apple'') into a semantic vector using CLIP’s text encoder. We then compute the similarity between this vector and each item of SVB $\textbf{B}_{s}$, selecting the patch with the highest similarity in each view. This allows us to obtain the 2D localization $L_{2D}$ of the target object in each view. This step can be formulated as:
  \begin{equation}
    L^{(i)}_{2D} = \operatorname{position}(\arg\max_j \mathcal{S}(\mathcal{E}_{text}(Q), \textbf{B}^{(i)}_{s}))
  \end{equation}
where $\mathcal{E}_{text}$ denotes text encoder, $\mathcal{S}(\cdot,\cdot)$ denotes the cosine similarity function, and $L^{(i)}_{2D} \in \mathbb{R}^{2}$ represents the 2D coordinates of the target object in the $i$-th view.
% 首先我们用CLIP的text encoder将文本查询编码成一个语义向量，然后与多视图的语义向量库进行相似度计算，选出每个view的下相似度最高的patch，这样便可以得到每个view下目标物体的二维定位。
% 再结合深度信息，我们可以将二维定位映射到三维空间中，从而得到目标物体的三维位置。
By incorporating depth information, we can map the 2D location $L^{(i)}_{2D}$ to 3D space, thereby obtaining the 3D position $L^{(i)}_{3D}$ of the target object. This step can be formulated as:
  \begin{equation}
    L^{(i)}_{3D} = \mathcal{P}^{-1}(L^{(i)}_{2D}, \textbf{B}^{(i)}_{d}, c_i)
  \end{equation}
where $\mathcal{P}^{-1}(\cdot,\cdot,\cdot)$ denotes the back-projection function that maps 2D coordinates to 3D space using the depth map and camera parameters, and $L^{(i)}_{3D} \in \mathbb{R}^{3}$ represents the 3D coordinates of the target object in the $i$-th view. Since each view can yield a 3D position and CLIP itself exhibits instability, erroneous retrieval results may occur. Therefore, we design a \textit{Multi-view Stereo Voting} strategy: we evaluate the Euclidean distances among the 3D positions obtained from each view, and the location indicated by the majority of views is regarded as the final 3D position $L_{3D} \in \mathbb{R}^{3}$.
% 由于每个view都能得到一个三维位置，并且CLIP本身具有不稳定性，会存在错误的检索结果。因此，我们设计了一种多视角投票策略：将每个视角得到的三维位置进行欧式距离判断，更多的视角指向的位置被认为是最终的三维位置。

\subsection{(C) Online Segmentation}
\label{ssec:seg}
As shown in Fig.~\ref{fig:framework} (C), this stage is for segmenting the target Gaussians based on the 3D position $L_{3D}$. First, we render a target-centered bird's-eye view (BEV) $v_{BEV}$ based on $L_{3D}$.
%首先，我们通过一直的目标三维位置，渲染出target-centered BEV。
Then, we perform point-driven segmentation that use points as prompts in the BEV view to obtain the mask $m_{BEV}$ of the target object, as:
% 然后，我们在BEV视图中进行目标分割，得到目标的2D掩码。
\begin{equation}
  m_{BEV} = \operatorname{seg}(\operatorname{render}(c_{BEV};\textbf{G}), L^{BEV}_{2D})
\end{equation}
where $L^{BEV}_{2D} = \mathcal{P}(L_{3D}, c_{BEV})$ denotes the 2D projection of the 3D point $L_{3D}$ onto the BEV plane, and $c_{BEV} = camera(L_{3D};\vec{u},\bar{r})$ denotes the camera pose that can be used for BEV rendering and $\{\vec{u}, \bar{r}\}$ represents the up vector and the height. By applying \textit{Frustum Filtering} in the BEV view, we can obtain a coarse grounding result.
% 通过BEV视图下的Frustum Filtering，我们可以进行一个圆柱形的选取，得到一个粗略的定位结果。

% \subsubsection{Frustum Filtering}
% \label{sssec:frustum filtering}
\noindent\textbf{Frustum Filtering (FF)} utilizes a 2D mask to filter 3D Gaussians through a mask-shaped frustum. Specifically, we project the positions of all Gaussian primitives onto the plane of the 2D mask; those that fall outside the mask are considered background, while those within the mask are identified as target Gaussians. This Frustum Filtering (FF) on each Gaussian can be formulated as:
\begin{equation}
\textbf{I}^{(i)}_{tar} = \mathcal{F}(\textbf{G}^{(i)})=
\begin{cases}
1, & \text{if } project(\textbf{G}^{(i)};c_{m}) > 0 \\
0, & \text{otherwise}
\end{cases}
\end{equation}
where $\mathcal{F}(\cdot)$ denotes FF, $\textbf{I}^{(i)}_{tar} \in \mathbb{R}^{1}$ denotes the flag of the $i$-th Gaussian primitive that 1 denotes target and 0 denotes background, and $c_{m}$ represents the camera parameters of plane of the 2D mask. With FF, we can quickly exclude Gaussian primitives that do not belong to the target object, thereby obtaining a coarse grounding result $\textbf{G}^{(\textbf{I}_{tar})}$ as:
% 我们能快速排除不属于目标物体的高斯基元，从而得到一个粗略的定位结果。
\begin{equation}
  \textbf{G}^{(\textbf{I}_{tar})} = \mathcal{F}(m_{2D};\textbf{G}, c_{plane})
\end{equation}
where $\textbf{G}^{(\textbf{I}_{tar})} \in \mathbb{R}^{S \times 59}$ denotes the set of Gaussian primitives that are identified as targets, and $S$ is the number of target Gaussians. $m_{2D}$ denotes the 2D mask used for filtering.

% \subsubsection{Surrounding Multi-view Frustum Intersection (SMFI)}
% \label{sssec:frustum intersection}
% 首先，利用3D Location生成k个环绕的虚拟相机，虚拟相机与target的距离可以在BEV上进行选择，一般为3倍的物体宽度。虚拟相机的俯仰角我们采用固定的30度。
\noindent\textbf{Surrounding Multi-view Frustum Intersection (SMFI)} refines the grounding results. First, we generate $k$ (typically $k=4$) surrounding virtual cameras $\textbf{C}_{vir}$ based on the 3D location $L_{3D}$. The distance $d_{vir}$ between each virtual camera and the target can be selected in the BEV, typically set to $3\times$ the target's width. The pitch angle $\theta_{vir}$ of the virtual cameras is fixed at $30^\circ$. This step can be formulated as:
\begin{equation}
  \textbf{C}^{(i)}_{vir} = Camera(\phi^{(i)};L_{3D}, d_{vir},\theta_{vir},\vec{u}), i=1,2,\ldots,k
\end{equation}
% 我们将coarse target Gaussians在虚拟相机上渲染出视图，然后把各自的视图喂入一个text-driven segmentation方法（本文选用的是Grounding DINO）。在得到每个视角下的2D掩码后，我们通过Frustum Filtering得到每个视角下的target Gaussians。我们并行处理每个视角，所以这一步可以做到实时。最后，我们将所有视角下的target Gaussians进行交集，得到最终的精细定位结果。
where $\phi^{(i)}$ denotes the yaw angle of the $i$-th virtual camera, which is uniformly sampled from $[0, 360^\circ)$, and $\vec{u}$ represents the up vector of the camera. Then, we render views of the coarse target Gaussians $\textbf{G}^{(\textbf{I}_{tar})}$ from each virtual camera, and feed these views into a text-driven segmenter (such as Grounded-SAM~\cite{liu2023grounding}). After obtaining the 2D mask for each view, we apply FF in each plane as:
\begin{equation}
  \tilde{\textbf{G}}^{(\textbf{I}_{tar})(i)} = \mathcal{F}(\operatorname{seg}(v_i,Q);\textbf{G}^{(\textbf{I}_{tar})}, c_{i})
\end{equation}

\begin{equation}
  \textbf{G}^{(\textbf{I}_{tar})} = \tilde{\textbf{G}}^{(\textbf{I}_{tar})(i)} \cap \ldots \cap \tilde{\textbf{G}}^{(\textbf{I}_{tar})(k)}
\end{equation}
where $\textbf{G}^{(\textbf{I}_{tar})} \in \mathbb{R}^{S \times 59}$ denotes the set of Gaussian primitives that are identified as the target, and $S$ is the number of target Gaussians.

\section{Experiments}

\begin{figure*}[htb]
  \centering
  \includegraphics[width=1\textwidth]{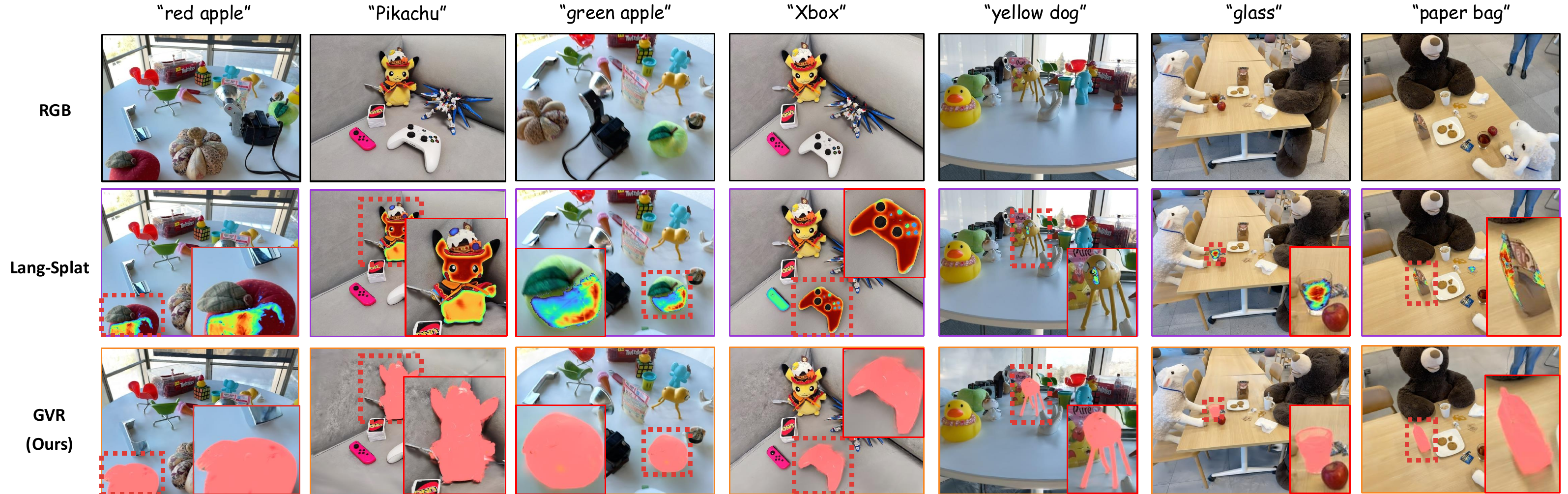}
    \vspace{-0.3in}
  \caption{\textbf{Qualitative comparison} between LangSplat~\cite{qin2024langsplat} and our GVR. Our method generates more precise and complete segmentation masks for the target objects described by the text queries. We visualize our results by dying Gaussians to red and splatting them on the camera plane.}
  \label{fig:qualitative}
  \vspace{-0.1in}
\end{figure*}

\definecolor{softred}{RGB}{255,178,178}
\definecolor{softorange}{RGB}{255,218,179}
\definecolor{softyellow}{RGB}{255,254,191}

\begin{table*}[thbp]
  \renewcommand\arraystretch{1}
  \tabcolsep=2pt
  \centering
  \caption{\textbf{Quantitative comparison with baselines and ablation study on LERF-Mask and 3D-OVS.} Metrics marked with $\uparrow$ indicate higher is better. \colorbox{softred}{1st}\colorbox{softorange}{2nd}\colorbox{softyellow}{3rd} denote the top three results. The ablation section reports the drop in performance compared to the full model.} 
  \label{tab:quant}
  \resizebox{\textwidth}{!}{%
  \begin{tabular}{c|cc|cc|cc|cc|cc|cccccc}
    \toprule
    \multirow{4}{*}{Methods} & \multicolumn{10}{c|}{\makecell{LERF-Mask}} & \multicolumn{6}{c}{\makecell{3D-OVS}} \\ \cline{2-17}
    & \multicolumn{2}{c|}{\makecell{Ramen}} & \multicolumn{2}{c|}{\makecell{Figurines}} & \multicolumn{2}{c|}{\makecell{Teatime}} & \multicolumn{2}{c|}{\makecell{Kitchen}} & \multicolumn{2}{c|}{\makecell{Overall}} & \multicolumn{6}{c}{\makecell{All}}\\
    & Acc$\uparrow$ & mIoU$\uparrow$ & Acc$\uparrow$ & mIoU$\uparrow$ & Acc$\uparrow$ & mIoU$\uparrow$ & Acc$\uparrow$ & mIoU$\uparrow$ & Acc$\uparrow$ & mIoU$\uparrow$ & bed & bench & room & sofa & lawn & \textbf{overall}\\
    \midrule
    LSeg~\cite{li2022language}  &14.1 &7.0 &8.9 &7.6 &33.9 &21.7 &27.3 &29.9 &21.1 &16.6 &56.0 &6.0 &19.2 &4.5 &17.5 &20.6 \\
    LERF~\cite{kerr2023lerf} &62.0 &28.2 &75.0 &38.6 &84.8 &45.0 &72.7 &37.9 &73.6 &37.4 &73.5 &53.2 &46.6 &27 &73.7 &54.8 \\
    LangSplat~\cite{qin2024langsplat} 
    &\cellcolor{softyellow}73.2 &\cellcolor{softyellow}51.2 
    &\cellcolor{softorange}80.4 &\cellcolor{softyellow}44.7 
    &\cellcolor{softorange}88.1 &\cellcolor{softorange}65.1 
    &\cellcolor{softorange}95.5 &\cellcolor{softyellow}44.5 
    &\cellcolor{softorange}84.3 &\cellcolor{softorange}51.4 
    &\cellcolor{softorange}92.5 &\cellcolor{softyellow}94.2 
    &\cellcolor{softyellow}94.1 &\cellcolor{softyellow}90.0 
    &\cellcolor{softyellow}96.1 &\cellcolor{softyellow}93.4 \\
    ReasonGrounder~\cite{liu2025reasongrounder} 
    &\cellcolor{softorange}78.5 &\cellcolor{softorange}53.4 
    &\cellcolor{softyellow}82.4 &\cellcolor{softorange}49.6 
    &\cellcolor{softyellow}89.7 &\cellcolor{softyellow}68.2 
    &\cellcolor{softred}\textbf{96.2} &\cellcolor{softorange}49.3 
    &\cellcolor{softyellow}86.7 &\cellcolor{softyellow}55.1 
    &\cellcolor{softyellow}93.3 &\cellcolor{softorange}96.6 
    &\cellcolor{softorange}94.5 &\cellcolor{softorange}91.7 
    &\cellcolor{softorange}97.3 &\cellcolor{softorange}94.7 \\
    Ours
    &\cellcolor{softred}\textbf{79.0}&\cellcolor{softred}\textbf{54.1}
    &\cellcolor{softred}\textbf{85.3}&\cellcolor{softred}\textbf{51.2}
    &\cellcolor{softred}\textbf{90.0} &\cellcolor{softred}\textbf{69.5} 
    &\cellcolor{softorange}95.8 &\cellcolor{softred}\textbf{50.1} 
    &\cellcolor{softred}\textbf{87.5} &\cellcolor{softred}\textbf{56.2} 
    &\cellcolor{softred}\textbf{93.1} &\cellcolor{softorange}96.6 
    &\cellcolor{softred}\textbf{95.3} &\cellcolor{softred}\textbf{94.3} 
    &\cellcolor{softred}\textbf{97.8} &\cellcolor{softred}\textbf{95.4} \\
    \midrule
    w/o RFL & -22.5 & -18.2 & -18.3 & -25.4 & -29.8 & -37.9 & -30.2 & -26.1 & -25.2 & -26.9 & -13.7 & -18.4 & -12.1 & -21.2 & -4.2 & -13.9\\
    w/o SMFI & -8.7 & -13.5 & -7.9 & -12.6 & -6.5 & -14.2 & -7.1 & -13.8 & -7.6 & -13.5 & -2.3 & -6.7 & -1.5 & -2.3 & -3.5 & -3.3\\
    w/o RFL \& SMFI & -23.6 & -19.3 & -20.7 & -27.1 & -30.8 & -38.6 & -33.2 & -27.8 & -26.6 & -18.2 & -15.9 & -20.2 & -13.4 & -23.1 & -5.1 & -15.5\\
    \bottomrule
  \end{tabular}
  }
  \vspace{-0.1in}
\end{table*}

% 我们从定性、定量两个方面对GVR进行评估。采用3DVG中最常用的两个数据集LERF-mask和3D-OVS。在定性分析上，我们采用定位准确率Accuracy和Mean IOU两个指标进行比较。

% For quantitative analysis, we report two key metrics: mean Intersection over Union (IoU) and Localization Accuracy (Acc). Following the protocol of LERF~\cite{kerr2023lerf}, a prediction is considered accurate if the point with the highest predicted relevance falls within the ground-truth bounding box. For qualitative analysis, we present visual comparisons of the segmentation results.

\textbf{Dataset}. We evaluate our method on two standard 3DVG benchmarks: LERF-Mask~\cite{kerr2023lerf} and 3D-OVS~\cite{liu2023weakly}.
The performance of GVR is evaluated using two main metrics: Localization Accuracy (Acc) and Intersection over Union (IoU).

\noindent\textbf{Implementation Details}. Our method utilizes three visual foundation models: SAM2~\cite{kirillov2023segment}, CLIP~\cite{radford2021learning}, and Grounding DINO~\cite{liu2023grounding}. In our experimental setting: for CLIP, we use OpenCLIP ViT-B/16 model; for SAM2, we use the ViT-H model; and for Grounding DINO, we use the DINO-SwinB model. The number of surrounding virtual cameras $k$ is set to 4, the distance $d_{vir}$ is set to 3 times the target's width, and the pitch angle $\theta_{vir}$ is fixed at $30^\circ$. All experiments are conducted on a single NVIDIA RTX 4090 GPU with 24GB memory.
%我们的方法采用了三个视觉基础模型包括SAM2，CLIP和Grounding-DINO。SAM2主要是将视图处理成object-level patches, CLIP用于提取文本特征，而Grounding-DINO则用于文本到对象的二维定位。

\noindent\textbf{Quantitative analysis}. As shown in Table~\ref{tab:quant}, our GVR achieves the best overall results on both LERF-Mask and 3D-OVS datasets. Specifically, GVR outperforms previous SOTA methods in both Accuracy and IoU, achieving 87.5\% Accuracy and 56.2\% IoU on LERF-Mask, and 95.4\% overall on 3D-OVS. These results demonstrate the effectiveness and generalization ability of our approach for zero-shot 3DVG.
% 如表所示，我们的GVR在LERF-Mask数据集上比previous SOTA ReasonGrounder有这更好的结果。在Accuracy上，GVR领先0.8%，在IOU上，GVR领先1.1%。

\noindent\textbf{Qualitative analysis}. As shown in Fig. ~\ref{fig:qualitative}, 7 queries are conducted on 3 scenes from LERF-Mask and 3D-OVS. Unlike LangSplat, which often localizes only part of the object due to its pixel-wise lookup, GVR operates on Gaussians and produces more accurate and complete masks. Notably, since ReasonGrounder isn't open-sourced, our qualitative analysis only compares with the second-best method, LangSplat.
% 我们分别在LERF-Mask和3D-OVS数据集中挑了三个场景，做了 6 个物体的3DVG。
% 由于LangSplat是pixel-wise的lookup，所以给定的文本查询通常只能定位到物体的一部分，而GVR则能完整地将物体分割出来。
\begin{table}[t]
\centering
  \vspace{-0.1in}
\caption{\textbf{Performance Comparison on Figurines scene.} Preparation time means the time for a method ready for queries. Query speed means the time for a single query.}
\label{tab:performance}
\begin{tabular}{lcc}
\hline
Method & Preparation Time & Query speed (s) \\
\hline
LangSplat & 1h 30min & 2.7 \\
ReasonGrounder & 48 min & 0.6 \\
\cellcolor{softred}\textbf{GVR (Ours)} & \cellcolor{softred}\textbf{37s} & \cellcolor{softred}\textbf{0.25} \\
\hline
\end{tabular}
\vspace{-0.2in}
\end{table}

\noindent\textbf{Performance Analysis}. As shown in Table~\ref{tab:performance}, our GVR significantly reduces both preparation time and query speed compared to previous methods. Unlike prior approaches that require time-consuming per-scene training, GVR only needs to build the knowledge books, enabling real-time querying.

% \subsection{Ablation Study}
% \label{ssec:ablation}
\noindent\textbf{Ablation on RFL}. We analyze the impact of the Retrieval for Localizing (RFL) module. If we do not use RFL, we directly use the text-driven segmentation on the BEV view to obtain the 2D mask for Frustum Filtering. As shown in Table~\ref{tab:quant}, without RFL, the performance drops significantly (from 87.5 to 62.3 in Acc and from 56.2 to 29.3 in mIoU), indicating that RFL is crucial for accurately localizing the target object and providing reliable clues for subsequent segmentation.

\noindent\textbf{Ablation on SMFI.} Without SMFI, only the coarse segmentation from the BEV frustum filtering is used. As shown in Table~\ref{tab:quant}, removing SMFI leads to a notable performance drop (from 87.5 to 79.9 in Acc and from 56.2 to 42.7 in mIoU), demonstrating that SMFI is effective in refining coarse results and improving the precision of target Gaussian segmentation.

\vspace{-0.1cm}
\section{Conclusions and Discussions}
\label{sec:conclusions}
We introduce GVR, a novel zero-shot 3DVG framework for 3DGS. By reformulating the 3DVG task as a 2D retrieval problem, GVR effectively leverages existing 2D visual foundation models to achieve high-quality localization without the need for 3D annotated data or per-scene training. Our extensive experiments demonstrate that GVR not only achieves state-of-the-art performance on standard benchmarks but also significantly reduces training time and query latency.

\vfill\pagebreak

\bibliographystyle{IEEEbib}
\bibliography{refs}

\begin{thebibliography}{10}

\bibitem{kerbl20233d}
Bernhard Kerbl, Georgios Kopanas, Thomas Leimk{\"u}hler, and George Drettakis,
\newblock ``3d gaussian splatting for real-time radiance field rendering.,''
\newblock {\em ACM Trans. Graph.}, vol. 42, no. 4, pp. 139--1, 2023.

\bibitem{GauGroup}
Mingqiao Ye, Martin Danelljan, Fisher Yu, and Lei Ke,
\newblock ``Gaussian grouping: Segment and edit anything in 3d scenes,''
\newblock in {\em European Conference on Computer Vision}. Springer, 2024, pp. 162--179.

\bibitem{choi2024click}
Seokhun Choi, Hyeonseop Song, Jaechul Kim, Taehyeong Kim, and Hoseok Do,
\newblock ``Click-gaussian: Interactive segmentation to any 3d gaussians,''
\newblock in {\em European Conference on Computer Vision}. Springer, 2024, pp. 289--305.

\bibitem{gaussianeditor}
Yiwen Chen, Zilong Chen, Chi Zhang, Feng Wang, Xiaofeng Yang, Yikai Wang, Zhongang Cai, Lei Yang, Huaping Liu, and Guosheng Lin,
\newblock ``Gaussianeditor: Swift and controllable 3d editing with gaussian splatting,''
\newblock in {\em Proceedings of the IEEE/CVF conference on computer vision and pattern recognition}, 2024, pp. 21476--21485.

\bibitem{SAGA}
Jiazhong Cen, Jiemin Fang, Chen Yang, Lingxi Xie, Xiaopeng Zhang, Wei Shen, and Qi~Tian,
\newblock ``Segment any 3d gaussians,''
\newblock {\em arXiv preprint arXiv:2312.00860}, 2023.

\bibitem{SAGD}
Xu~Hu, Yuxi Wang, Lue Fan, Junsong Fan, Junran Peng, Zhen Lei, Qing Li, and Zhaoxiang Zhang,
\newblock ``Sagd: Boundary-enhanced segment anything in 3d gaussian via gaussian decomposition,''
\newblock {\em arXiv preprint arXiv:2401.17857}, 2024.

\bibitem{shen2024flashsplat}
Qiuhong Shen, Xingyi Yang, and Xinchao Wang,
\newblock ``Flashsplat: 2d to 3d gaussian splatting segmentation solved optimally,''
\newblock in {\em European Conference on Computer Vision}. Springer, 2024, pp. 456--472.

\bibitem{zhu2025rethinking}
Runsong Zhu, Shi Qiu, Zhengzhe Liu, Ka-Hei Hui, Qianyi Wu, Pheng-Ann Heng, and Chi-Wing Fu,
\newblock ``Rethinking end-to-end 2d to 3d scene segmentation in gaussian splatting,''
\newblock {\em arXiv preprint arXiv:2503.14029}, 2025.

\bibitem{liu2025reasongrounder}
Zhenyang Liu, Yikai Wang, Sixiao Zheng, Tongying Pan, Longfei Liang, Yanwei Fu, and Xiangyang Xue,
\newblock ``Reasongrounder: Lvlm-guided hierarchical feature splatting for open-vocabulary 3d visual grounding and reasoning,''
\newblock in {\em Proceedings of the Computer Vision and Pattern Recognition Conference}, 2025, pp. 3718--3727.

\bibitem{li2025scenesplat}
Yue Li, Qi~Ma, Runyi Yang, Huapeng Li, Mengjiao Ma, Bin Ren, Nikola Popovic, Nicu Sebe, Ender Konukoglu, Theo Gevers, et~al.,
\newblock ``Scenesplat: Gaussian splatting-based scene understanding with vision-language pretraining,''
\newblock {\em arXiv preprint arXiv:2503.18052}, 2025.

\bibitem{qin2024langsplat}
Minghan Qin, Wanhua Li, Jiawei Zhou, Haoqian Wang, and Hanspeter Pfister,
\newblock ``Langsplat: 3d language gaussian splatting,''
\newblock in {\em Proceedings of the IEEE/CVF Conference on Computer Vision and Pattern Recognition}, 2024, pp. 20051--20060.

\bibitem{kirillov2023segment}
Alexander Kirillov, Eric Mintun, Nikhila Ravi, Hanzi Mao, Chloe Rolland, Laura Gustafson, Tete Xiao, Spencer Whitehead, Alexander~C Berg, Wan-Yen Lo, et~al.,
\newblock ``Segment anything,''
\newblock in {\em Proceedings of the IEEE/CVF international conference on computer vision}, 2023, pp. 4015--4026.

\bibitem{radford2021learning}
Alec Radford, Jong~Wook Kim, Chris Hallacy, Aditya Ramesh, Gabriel Goh, Sandhini Agarwal, Girish Sastry, Amanda Askell, Pamela Mishkin, Jack Clark, et~al.,
\newblock ``Learning transferable visual models from natural language supervision,''
\newblock in {\em International conference on machine learning}. PmLR, 2021, pp. 8748--8763.

\bibitem{liu2023grounding}
Shilong Liu, Zhaoyang Zeng, Tianhe Ren, Xuefei Ning, Zhiyang Dou, Zhiqiang Shen, Xiangyu Zhang, Yilun Chen, Yifei Huang, Yixuan Wei, Yanjie Han, Yutong Bai, Hongyang Li, Zehuan Yuan, and Jifeng Dai,
\newblock ``Grounding dino: Marrying dino with grounded pre-training for open-set object detection,''
\newblock in {\em ICCV}, 2023.

\bibitem{li2022language}
Boyi Li, Kilian~Q Weinberger, Serge Belongie, Vladlen Koltun, and Ren{\'e} Ranftl,
\newblock ``Language-driven semantic segmentation,''
\newblock {\em arXiv preprint arXiv:2201.03546}, 2022.

\bibitem{kerr2023lerf}
Justin Kerr, Chung~Min Kim, Ken Goldberg, Angjoo Kanazawa, and Matthew Tancik,
\newblock ``Lerf: Language embedded radiance fields,''
\newblock in {\em Proceedings of the IEEE/CVF international conference on computer vision}, 2023, pp. 19729--19739.

\bibitem{liu2023weakly}
Kunhao Liu, Fangneng Zhan, Jiahui Zhang, Muyu Xu, Yingchen Yu, Abdulmotaleb El~Saddik, Christian Theobalt, Eric Xing, and Shijian Lu,
\newblock ``Weakly supervised 3d open-vocabulary segmentation,''
\newblock {\em Advances in Neural Information Processing Systems}, vol. 36, pp. 53433--53456, 2023.

\end{thebibliography}
\end{document}